\documentclass[sigconf,screen]{acmart}

\usepackage{multirow}
\usepackage[ruled,vlined]{algorithm2e}

\renewcommand\footnotetextcopyrightpermission[1]{} 
\AtBeginDocument{%
  \providecommand\BibTeX{{%
    \normalfont B\kern-0.5em{\scshape i\kern-0.25em b}\kern-0.8em\TeX}}}

\setcopyright{acmcopyright}
\copyrightyear{2021}
\acmYear{2021}
\acmDOI{10.1145/1122445.1122456}

\begin{document}

\title{Structure Enhanced Graph Neural Networks for Link Prediction}


\author{Baole Ai}
\affiliation{%
  \institution{Alibaba Group}
  \city{Hangzhou}
  \country{China}}
\email{baole.abl@alibaba-inc.com}

\author{Zhou Qin}
\affiliation{%
  \institution{Alibaba Group}
  \city{Hangzhou}
  \country{China}}
\email{qinzhou.qinzhou@alibaba-inc.com}


\author{Wenting Shen}
\affiliation{%
  \institution{Alibaba Group}
  \city{Hangzhou}
  \country{China}}
\email{wenting.swt@alibaba-inc.com}

\author{Yong Li}
\affiliation{%
  \institution{Alibaba Group}
  \city{Hangzhou}
  \country{China}}
\email{jiufeng.ly@alibaba-inc.com}

\renewcommand{\shortauthors}{Ai and Zhao, et al.}

\begin{abstract}
Graph Neural Networks (GNNs) have shown promising results in various tasks, among which link prediction is an important one. GNN models usually follow a node-centric message passing procedure that aggregates the neighborhood information to the central node recursively. Following this paradigm, features of nodes are passed through edges without caring about where the nodes are located and which role they played. However, the neglected topological information is shown to be valuable for link prediction tasks. In this paper, we propose Structure Enhanced Graph neural network (SEG) for link prediction. SEG introduces the path labeling method to capture surrounding topological information of target nodes and then incorporates the structure into an ordinary GNN model. By jointly training the structure encoder and deep GNN model, SEG fuses topological structures and node features to take full advantage of graph information. Experiments on the OGB link prediction datasets demonstrate that SEG achieves state-of-the-art results among all three public datasets.
\end{abstract}




\maketitle
\pagestyle{plain}

\section{Introduction}

\begin{figure*}[htbp]
  \centering
  \includegraphics[width=0.9\linewidth]{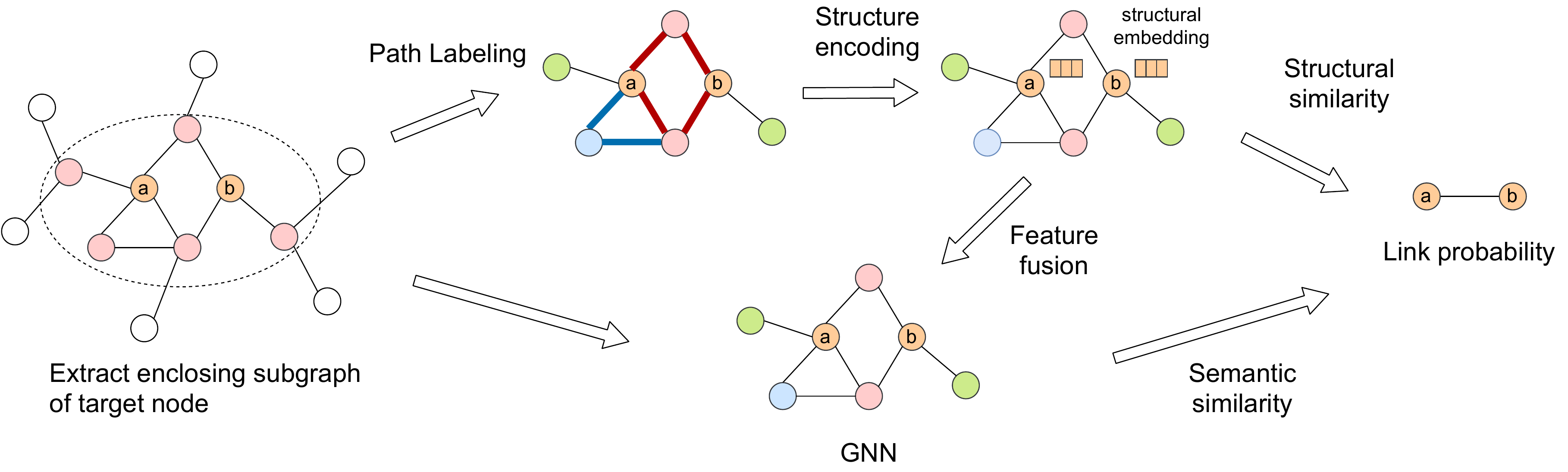}
  \caption{The SEG framework. For target nodes $a$ and $b$, SEG first extracts an enclosing subgraph around it, and then extracts paths and jointly trains an encoder and a GNN to learn both graph structure and the original node features for link prediction. Paths are indicated by red and blue bolded edges.}
  \Description{The SEG framework. For target nodes $a$ and $b$, SEG first extracts an enclosing subgraph around it, and then extracts paths and jointly trains an encoder and a GNN to learn both graph structure and the original node features for link prediction. Paths are indicated the red and blue bolded edges.}
  \label{fig:intro}
\end{figure*}
Link prediction, which is used to predict whether two nodes
are connected in a graph, is widely applied in various fields, such as recommendation, knowledge graph, and social network. 
The traditional approaches mainly focus on graph topology. Topology-based methods compute node similarity score as the likelihood of connections, such as Common Neighbors (CN)~\cite{newman2001clustering}, Jaccard measure~\cite{chowdhury2010introduction}, Adamic/Adar (AA) measure~\cite{ADAMIC2003211}, and Katz measure~\cite{katz1953new}. 
Graph embedding approaches are currently popular to resolve link prediction problems. Representative methods such as DeepWalk~\cite{perozzi2014deepwalk}, LINE~\cite{tang2015line} and Node2vec~\cite{grover2016node2vec} are trying to learn parameter-free low-dimensional node embeddings from the observed graph nodes and then use these embeddings to predict links. These approaches potentially belong to transductive learning and therefore cannot predict unseen nodes.
Recently, graph neural networks (GNNs) has shown impressive performance on link prediction~\cite{kipf2016semi, hamilton2017inductive, zhang2018end}. GNNs generally follow a message passing schema to aggregate neighborhood vectors recursively and have been proved to be as powerful as the Weisfeiler-Lehman graph isomorphism test~\cite{xu2018how}. 
However, only features of nodes are passed during the message passing procedure, topological information of nodes is not explicitly considered. These topological features are shown to be useful in topology-based methods. 
Moreover, on some datasets, topology-based methods such as CN are more effective than GNN methods such as GCN~\cite{kipf2016semi} and GraphSAEG~\cite{hamilton2017inductive}.
Based on these observations, scholars began to explicitly extract topological information as an extra input of GNNs~\cite{zhang2018link, li2020distance}. The SEAL~\cite{zhang2018link} method first extracts an enclosing subgraph around
the target link and then labels nodes in the subgraph based on their different location. Then a GNN framework named DGCNN~\cite{zhang2018end} is applied to predict the existence of the target link.  
These works simply use node labeling tricks and directly concatenate structural features with original node features as the input of GNNs.
However, which structural features are useful and how to incorporate them with GNNs is still underexplored.

In this work, we attempt to resolve two problems: 1. How to design and encode structural features. 2. How to incorporate structural features into GNNs for link prediction. 
For the first problem, motivated by topology-based methods, we propose a novel node position labeling method called \textbf{Path Labeling(PL)} to extract nodes' structural features. Instead of directly using these structural features as the input of a GNN model like SEAL~\cite{zhang2018link}, we design an encoder to transform structural features to structural embeddings.
For the second problem, we propose a novel \textbf{Structure Enhanced Graph neural network} (SEG) to integrate structural embeddings into GNNs.
Given structural embeddings mentioned above, SEG uses a feature fusion module to map the structural embeddings and the original node features into the same embedding space. Fused result are then used as the initial input of a GNN model. Through jointly training the structure encoder and GNN, SEG can learn both structure and attribute information of nodes, thus optimizes the usage of graph information for link prediction. The pipeline of SEG is illustrated in Figure~\ref{fig:intro}.
To predict the link between two target nodes $a$ and $b$, 
we first extract an enclosing 1-hop subgraph around them. Then we label each node according to their position role in the subgraph to get node structural features, and apply a structure encoder to generate structural embeddings for each node. These structural embeddings are fused with the node features as the input of a GNN model.
The output of the GNN model and correlation between structural embeddings of these two target nodes are jointed together to make the final prediction of whether the link exists.
Our contributions are summarized as follows: 
\begin{itemize}
    \item We propose a novel node position labeling method called Path Labeling(PL) to encode structural features of graph nodes.
    \item We propose an end-to-end joint training framework named SEG that learns both structural features and initial node features.
    \item We validate SEG on three OGB\footnote{https://ogb.stanford.edu/docs/linkprop/} link-prediction datasets and achieve the state-of-art result.
\end{itemize}

The rest of the paper is organized as follows. In Section~\ref{sec:related_work}, we will review some related works. Section~\ref{sec:pre} defines the problem and lists notations we use. We then introduce the proposed method Structure Enhanced Graph neural networks (SEG) in Section~\ref{sec:approach} and discuss experimental setup and insights in
Section~\ref{sec:exp}. Finally, the whole paper is concluded in Section~\ref{sec:conclusion}.

\section{Related Work} \label{sec:related_work}
In this section, we briefly survey related work in graph neural networks and link prediction.

\subsection{Graph Neural Networks}

Graphs are ubiquitous in the real world and have a wide range of applications in many fields, including social networks, biological networks, the co-authorship network and the World Wide Web. 
Over the past few years, Graph Neural Networks (GNNs) have achieved great success on many tasks, including node classification, link prediction, graph classification, and recommender systems\cite{kipf2016semi, hamilton2017inductive, zhang2018link, zhang2018end, ying2018graph}.
GNN uses deep learning methods to solve graph-related problems, which can be seen as an application of deep learning on graph data. One of the most popular methods of GNNs called Graph Convolutional Networks (GCNs). GCNs originated with the work of Bruna et al.~\cite{bruna2013spectral}, which develops a version of graph convolutions based on spectral graph theory. 
Bruna et al.~\cite{bruna2013spectral} generalize the convolution
operation from Euclidean data to graph data by using the Fourier basis of a given graph. And then Defferrard et al.~\cite{defferrard2016convolutional} propose ChebNet, which utilizes Chebyshev polynomials as the filter to simplify the calculation of the convolution. Kipf et al.~\cite{kipf2016semi} further simplify ChebNet by using its first-order approximation which is the well-known GCN model. 
Some approaches define graph convolutions in the spatial domain~\cite{hamilton2017inductive, velivckovic2017graph}. 
Hamilton et al.~\cite{hamilton2017inductive} introduce GraphSAGE which first samples a fixed-sized of neighborhood nodes and then aggregates neighbor information using different strategies. DGCNN~\cite{zhang2018end} proposes a SortPooling layer that
sorts graph vertices in a consistent order and then uses a traditional 1-D convolutional neural network for graph classification.
Veličković et al.~\cite{velivckovic2017graph} proposes the graph attention network (GAT) model which uses the attention mechanism to assign different weights to different neighborhood nodes when aggregating information.  

More recently, there has been a lot of works try to improve the training efficiency of GNNs. FastGCN~\cite{chen2018fastgcn} performs layer-wise sampling to solve the neighbor explosion problem of node-wise sampling methods like GraphSAGE~\cite{hamilton2017inductive}. ClusterGCN~\cite{cluster_gcn} proposes graph clustering based training method. GraphSAINT~\cite{zeng2019graphsaint} proposes the subgraph sampling based training method. 

\subsection{Link Prediction}
Link prediction is a fundamental problem for graph data. It focus on predicting the existence of a link between two target nodes~\cite{liben2007link}. Classical approaches for link prediction are topology-based heuristic methods. Topology-based methods use graph structural features to calculate nodes' similarity as the likelihood of whether they are connected. Some representative methods include Common Neighbors (CN)~\cite{newman2001clustering}, Jaccard measure~\cite{chowdhury2010introduction}, Adamic/Adar (AA) measure~\cite{ADAMIC2003211}, and Katz measure~\cite{katz1953new}. CN computes the number of common neighbors as similarity scores to predict the link between two nodes. Jaccard measure computes the relative number of neighbors in common. CN and Jaccard 
only use one-hop neighbors, while the AA method captures two-hop similarities instead. The Katz measure further uses longer paths to capture a larger range of local structures.

Other common approaches are graph embedding methods, which first embed the nodes into low-dimensional space and then calculate the similarity of the nodes' embeddings to predict the existence of a link. DeepWalk~\cite{perozzi2014deepwalk} first generates random walks and then uses Word2vec~\cite{mikolov2013efficient} to get the embeddings of nodes.  LINE~\cite{tang2015line} introduces concepts of first-order and second-order similarity into weighted graphs. Node2vec~\cite{grover2016node2vec} extend DeepWalk with a more sophisticated random walk strategy that combines the DFS and BFS. 

More recently, GNNs~\cite{kipf2016semi,hamilton2017inductive} have been used for link prediction. GNN-based methods typically use supervised learning to model the link prediction as a classification problem. Zhang and Chen~\cite{zhang2017weisfeiler} propose WLNM to extract the enclosing subgraph around the target link and then encode the subgraph as a vector for subsequent classification. Zhang and Chen~\cite{zhang2018link} propose SEAL which first labels the nodes in each enclosing subgraph according to their distances to the source and target nodes, and then applies DGCNN~\cite{zhang2018end} to learn a link representation for link prediction. Li et al.~\cite{li2020distance} introduces another node labeling trick similar to SEAL which directly uses the shortest distances to target nodes. You et al .~\cite{you2019position} propose Position-aware GNN (P-GNN)~\cite{you2019position}. P-GNN uses anchor nodes to introduce relative location information of nodes into GNNs.
The model proposed in this paper also uses the labeling trick introduced in SEAL~\cite{zhang2018link}. However, SEAL simply concatenates structural features with original features of the nodes as the input of GNNs, which will be improved in this paper.

\section{Preliminaries} \label{sec:pre}

We consider an undirected graph which can be represented as a graph $G=(V, E, A)$, where $V=\{v_1,v_2,...,v_N\}$ is the set of $N$ nodes, $E \subseteq |V| \times |V|$ is the set of edges, and $A$ is the adjacency matrix of the graph, where $A_{i,j} =1$ if $(i,j) \in{E}$ and $A_{i,j}=0$ otherwise. A graph can be associated with node features matrix $X=\{x_1,x_2,...,x_N\}$ , where $X \in{\mathbb{R}^{N \times D}}$ and $D$ is the dimension of node features.

The goal of link prediction is to determine whether there exists an edge $e_{i,j}$ between two given nodes $\{v_i, v_j\}$. It can be formulated as a classification problem on potential links $E_p$ given observed edges $E_o$ and observed node features $X_o$. 

General GNN algorithm follows a message passing strategy, where each node's features can be iteratively updated by aggregating its neighbors'. During each iteration, hidden vector of node $u$ is updated by aggregating information from it's neighborhood $N(u)$. The $k$-th iteration/layer hidden vector of node $u$ is calculated as follows:
\begin{equation}
\begin{split}
      h_u^{(k)} = COMBINE^{(k)}(&h_u^{(k-1)}, \\ &AGGREGATE^{(k)}({h_v^{(k-1)},
  \forall v \in N(u)})) \label{eq:agg}
\end{split}
\end{equation}
where $AGGREGATE^{(k)}$ is the function of neighborhood aggregation and $COMBINE^{(k)}$ represents the process of combining the aggregated neighborhood vectors and its own vector $h_u^{(k-1)}$. We initialize these embeddings using node features, i.e., $h_u^{(0)} = x_u, \forall u \in V$. After running $K$ iterations, we can get the final layer vector $h_u^{(K)}, \forall u \in V$ for each node. These vectors can be used in downstream tasks such as node classification and link prediction, etc.

\section{APPROACH} \label{sec:approach}

In this section, we will detail our Structure Enhanced Graph neural network (SEG) model. 
Most GNN models treat edges in a graph as message passing routers, while ignoring topological information beneath them. However, topology-based methods, such as AA and CN, make use of structural information directly to successfully predict links.

Inspired by these methods, we propose a novel structure encoding method to extract structural embeddings for each node. To effectively incorporate structural features into GNN, we design the SEG framework to jointly train a deep GNN and a structure encoder.

\subsection{Structure Encoding} \label{sec:cp}

\begin{figure}[h]
  \centering
  \includegraphics[width=\linewidth]{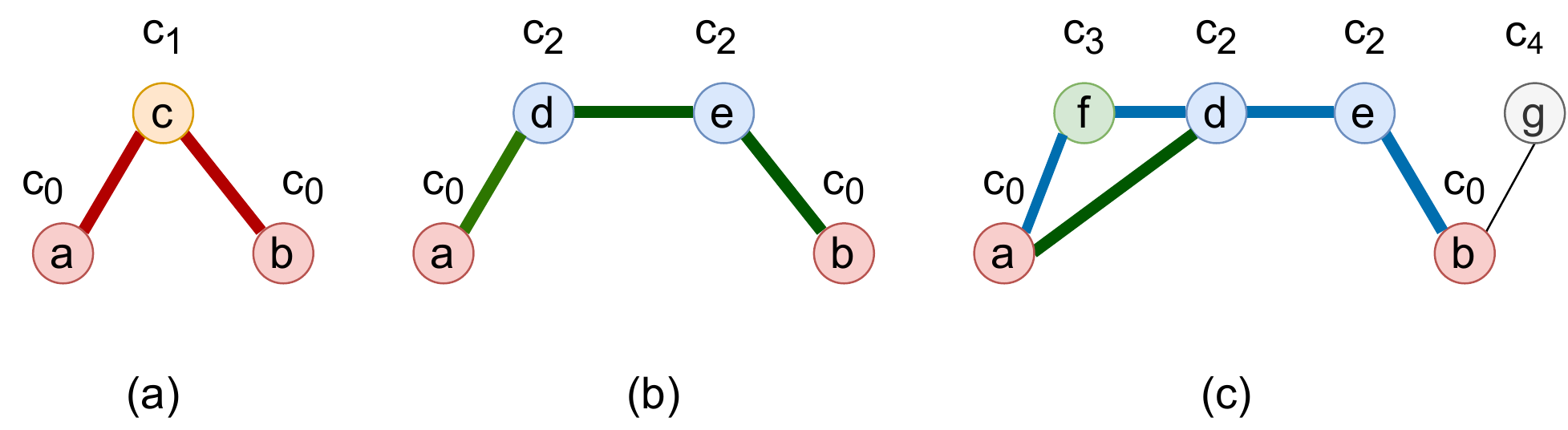}
  \caption{Examples of path labeling(PL) on 1-hop subgraph of nodes $a$ and $b$ . (a) is the path of length 2, (b) is the path of length 3, and (c) is the path of length 4. Different paths are drawn by different colors. Labels of nodes are marked above nodes and nodes labeled differently also are drawn by different colors.}
  \label{fig:cp}
\end{figure}

We first focus on the question that how to design
and encode structural features.
There are many local structural patterns around target nodes, such as common neighbors and triangles. 
To measure the correlation between two target nodes, we need to examine the topology of these nodes and edges connect them at the same time.
Graph structures like Common Neighbors (CN)~\cite{newman2001clustering}, Jaccard measure~\cite{chowdhury2010introduction}, Adamic/Adar (AA) measure~\cite{ADAMIC2003211}, and Katz measure~\cite{katz1953new} are shown to be useful for link prediction. These methods can be unified as follows:
\begin{equation}    
s(i,j) = \sum_{l=1}^{\infty }{f(l, N(i), N(j))\phi(p^l_{i,j})}\label{eq:topo} 
\end{equation}
where $s(i,j)$ denotes similarity between node $v_i$ and $v_j$, $p^l_{i,j}$ denotes number of length-$l$ paths between target nodes $v_i$ and $v_j$, and $N(i)$, $N(j)$ means neighbor nodes of $v_i$ and $v_j$, respectively. For example, suppose $f(l, N(i), N(j))=\alpha^l$ and $\phi(p_{i,j}^l)=p_{i,j}^l$, Equation~\ref{eq:topo} is the same as the Katz measure $\text{Katz}_{i,j}=\Sigma^\infty_{l=1}\alpha^l p_{i,j}^l$.
If $l = 1$, the equation is the same as neighbor based methods like CN and AA.

\begin{figure*}[htbp]
  \centering
  \includegraphics[width=0.7\linewidth]{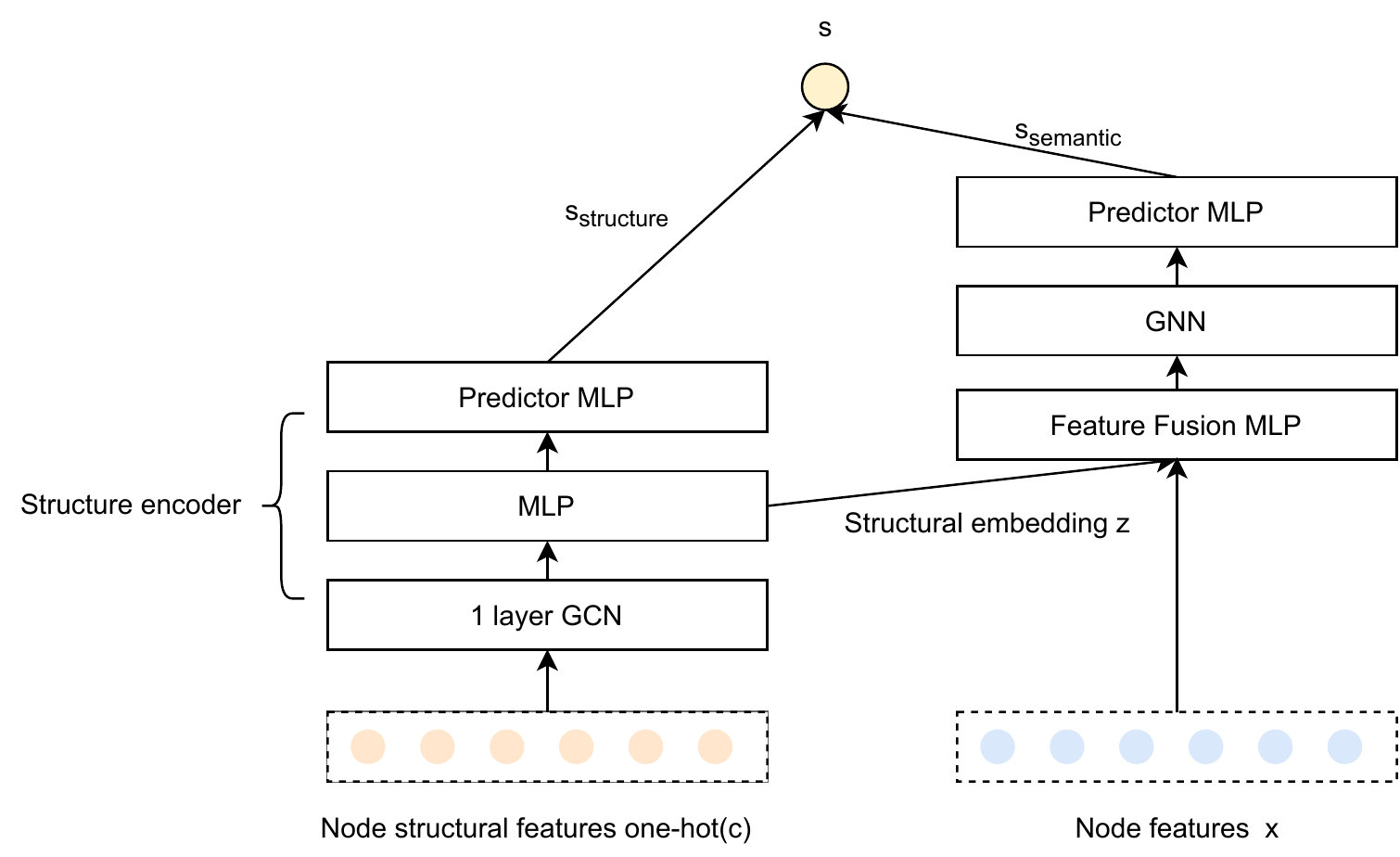}
  \caption{Architecture of SEG. It consists of two branches of neural networks, one is structure encoder to learn structural
  features and the other is a deep GNN model which uses fused features as input. These two types of embeddings are passed to Predictor MLPs respectively to get structural similarity score and high-level semantic similarity score, these two scores are jointed to be the final link existence probability.}
  \label{fig:arch}
\end{figure*}

Two key components of Equation~\ref{eq:topo} are paths of target nodes $p^l_{i,j}$ and their neighbor nodes $N(i), N(j)$. Since information of neighbors is already considered in GNNs, paths are especially important to be an extra input of GNN models. Here we propose path between two targets nodes as a kind of topological feature, defined as follows:
\begin{definition}[path]
A path $p_{i,j}$ between two target nodes $v_i$ and $v_j$ is a sequence of vertices and undirected edges starting with $v_i$ and ending with $v_j$ in which neither vertices nor edges are repeated.
\end{definition}

To numericalize these paths as usable topological features for nodes,
we propose a \textbf{Path Labeling}(PL) method to label each node $u$ with a discrete value $c_u$ based on the path it belongs to. Basically, 
nodes in different paths of different lengths are labeled with different discrete values. The fringe nodes that are not part of the path are labeled with a default value. For example, all 1-hop fringe nodes have an identical label value. The target nodes are also labeled with a default value.
For a node that appears in multiple paths at the same time, we choose the shortest path to label the node.

Figure~\ref{fig:cp} shows examples of path labeling in the subgraph which consists of 1-hop neighbors of two target nodes $a$ and $b$.
We label two target nodes $a$ and $b$ with a discrete value $c_0$; The node $c$ in (a) is the common neighbor of target nodes, and we label it with $c_1$; Nodes $d$ and $e$ in (b) are located on the path of length 3, we label them with $c_2$; Nodes $d$ and $e$ in (c) are located in both the path of length 3 (a-d-e-b) and length 4 (a-f-d-e-b), so we choose the path of length 3 to label $d$ and $e$ with value $c_2$, and label $f$ in (c) with $c_3$. Node $g$ is a fringe node, so we label it with a default value $c_4$.

Given these topological features extracted from paths of each pair of target node, we further propose a structure encoding method to learn representation vectors from them. The encoding method is defined as follows:
\begin{definition}[Structure Encoding]
Given structural features of node ${v_u}$ by path labeling method, we design a structure encoder as a mapping $c_u \to \mathbb{R}^m$, which transform structural features to structural embeddings, where $m$ is the dimension of representation vectors of nodes.
\end{definition}
This encoding process can be implemented with neural networks. In Algorithm~\ref{algo:cpe}, we introduce our structure encoding method. We first extract the paths $p_{i, j}$ and then generate structural features $c_u$ for each node $u\in G_S$ according to path labeling(PL). Then we use one-hot method to generate the initial structural embedding of node $u$ according to $c_u$, notes that $c_u$ is first truncated with a constant $\lambda$ to prevent too long paths. Finally a GCN layer and a MLP is applied to get the final node structural embeddings.
\begin{algorithm}[h]
\SetAlgoLined
\LinesNumbered
\KwIn{target nodes $i$, $j$; enclosing subgraph $G_S$}
\KwOut{node embedding $z$}
/* extracts the paths */ \\
$p_{i,j} \leftarrow (G_S, i, j)$ \;
/* generate node structural features*/ \\
$c_u \leftarrow \{p_{i,j}, G_S\}, \forall u \in G_S$ \;
/* generates one-hot embeddings*/ \\
$z_u^{(0)} \leftarrow$ one-hot$(min(c_u, \lambda)), \forall u \in G_S$ \;
/* encode with a GCN layer and a MLP */ \\
\For{$u \in G_S$}{
\emph{$z_u^{(1)} \leftarrow AGGREGATE(z_v^{(0)}, \forall v \in N(u))$} \;
}
$z_u \leftarrow MLP(z_u), \forall u \in G_S$
 \caption{structure encoding}
 \label{algo:cpe}
\end{algorithm}

In Algorithm~\ref{algo:cpe}, we use a GCN layer and MLP to fit $f$ and $\phi$ in Equation~\ref{eq:topo}. With appropriate parameters and enough layers, we hope the error of approximation can be ignored.


By the way, SEAL~\cite{zhang2018link} also uses labeling method to generate structural features. It's Double-Radius Node Labeling(DRNL) method labels each node according to its radius with respect to the two target nodes. The difference is that DRNL is much more strict than PL, because it distinguishes different nodes in the same length path. For example, given a path (a-b-c-d-e), where $a$ and $e$ are target nodes, the labeling result of DRNL is ($c_0$, $c_1$, $c_2$, $c_1$, $c_0$) and the result of PL is ($c_0$, $c_1$, $c_1$, $c_1$, $c_0$). In addition, we use a constant $\lambda$ to truncate too long paths, these difference prevent PL from overfitting thus more robust than DRNL.

\subsection{Feature Fusion}

To learn both node attributes and structural information, one straightforward way is to directly concatenate node features and structural features as the input of GNNs like SEAL~\cite{zhang2018link}. However, these two features express different meanings. The node attributes are usually constructed by semantic information, while the structural features are directly extracted from graph topology. How to learn both semantic information and structural information at the same time is also a challenging problem.

We design a framework to fuse these two features called SEG. The detail architecture is shown in Figure~\ref{fig:arch}. 
The SEG architecture contains two parts, one is a structure encoder and the other is a deep GNN. In the structure encoder, we start at encoding node structure features ($c_u$) with a single layer GCN. The reason for choosing one layer GCN is that the multi-hop information is already included in the path, so only one layer of aggregation is needed to update two target nodes we focus on. The output of the GCN layer are used as the input of a MLP to fit $f$ and $\phi$ in Equation~\ref{eq:topo}. We denote the output vectors $z_u$ of MLP as the structural embeddings of $v_u$. Another MLP is used to predict the structure similarity score $s_{structure}$ based on the Hadamard product of structural embeddings of target nodes $z_i$ and $z_j$.
\begin{equation}
    s_{structure} = MLP(z_i \circ z_j)
    \label{eq:ip} 
\end{equation}
For the deep GNN model, structural embedding $z_u$ and original node attributes $x_u$ are firstly fused by a MLP. The module of feature fusion is to map $x$ and $z$ into the same feature space. With fused features as the input, a deep GNN model can learn from both structural features and original node features. Here we choose a multi-layers GraphSAGE~\cite{hamilton2017inductive} as our backbone GNN model and use a SortPooling~\cite{zhang2018end} layer to get the final representation output of two targets nodes. Finally, we use a MLP to predict the high-level semantic similarity score, denoted as $s_{semantic}$. We combine the structure similarity score and semantic similarity score as the final link existence probability:
\begin{equation}
    s = s_{semantic} + s_{structure} 
    \label{eq:score} 
\end{equation}

\subsection{Training Algorithm}
In this section, we introduce the SEG training algorithm. The GNN updating procedure can be formed in Equation~\ref{eq:agg}. However, in practice, considering the efficiency of training, especially for real-world large graphs, we do not perform the above update process directly on the whole graph~\cite{hamilton2017inductive, cluster_gcn, aligraph}. Instead, we  extract a $k$-hop enclosing subgraph $G_S$ of two target nodes $i, j$ from the whole graph $G$. 
A $k$-hop enclosing subgraph is a graph consisting of $k$-hop neighbors of nodes $i$ and $j$. 
Then, we apply structure encoding described in Algorithm~\ref{algo:cpe} on $G_S$. Finally, two parts of SEG is jointly training using a sigmoid cross entropy loss,
\begin{equation}
    loss = \frac{1}{N}\sum_{t=1}^{N}{-y_t\log{s_t}-(1-y_t)log(1-s_t)} \label{eq:loss} 
\end{equation}
where $s_t$ refers to the probability score that the link $t$ is predicted to be true, $y_t$ is the label of the link $t$, and $N$ is the number of training edges. Algorithm~\ref{algo:train} details how to generate the final predict score.
During training, we randomly sample two nodes that are not connected as negative samples.

\begin{algorithm}[h]
\SetAlgoLined
\LinesNumbered
\KwIn{target edge $(i, j)$; input graph $G$;  node features $X$}
\KwOut{prediction score $s$}
/* extracts enclosing subgraph */ \\
$G_S \leftarrow G$ \;
$z_u \leftarrow Structural~Encoding(G_S, i, j), \forall u \in G_S$ \;
$s_{structure} \leftarrow MLP(z_i \circ z_j)$ \;
/* feature fusion */ \\
$\tilde{x}_u^{(0)} \leftarrow x_u^{(0)} + z_u, \forall u \in G_S$ \;
$\tilde{x}_u \leftarrow MLP(\tilde{x}_u^{(0)}), \forall u \in G_S$\;
$h_u^{(0)} \leftarrow \tilde{x}_u, \forall u \in G_S$ \;
/* GNN message passing*/ \\
\For{$k = 1,...,K$}{
\For{$u \in G$}{
\emph{$h_u^{(k)} \leftarrow$ Equation~\ref{eq:agg}} \;
}
}
$ h_G \leftarrow SortPool({h_u^{(k)} | u \in G_S, k=1,...,K}) $ \;
$s_{semantic} \leftarrow MLP(h_G)$ \;
$s \leftarrow s_{semantic} + s_{structure}$

\caption{SEG}
\label{algo:train}
\end{algorithm}

\section{Experiments} \label{sec:exp}
In this section, we conduct experiments on real-world datasets to
evaluate the effectiveness of SEG. In particular, we attempt to
answer the following questions:
\begin{itemize}
    \item Can SEG outperform existing baselines for link prediction on various benchmark datasets?
    \item How effective are structure encoding and joint training of SEG for link prediction?
    \item Is the Path Labeling(PL) more effective than the Double-Radius Node Labeling(DRNL) of SEAL?
\end{itemize}

\subsection{Experimental Setup}
\begin{table}
  \caption{Summary of ogbl-ppa, ogbl-collab and ogbl-citation2}
  \label{tab:data}
  \begin{tabular}{l|ccc}
    \toprule
	Dataset & \#Nodes & \#Edges & Feature Dimension \\
    \midrule
    ogbl-ppa & 576,289 & 30,326,273 & 58\\
    ogbl-collab & 235,868 & 1,285,465 & 128\\
    ogbl-citation2 & 2,927,963 & 30,561,187 & 128\\
    \bottomrule
  \end{tabular}
\end{table}

\begin{table*}[htb]
  \caption{Results for ogbl-ppa, ogbl-collab and ogbl-citation2}
  \label{tab:ogb}
  \begin{tabular}{l|ccc}
    \toprule
	\multirow{2}{*}{Method} & ogbl-ppa & ogbl-collab & ogbl-citation2 \\
	~ & Hits@100 & Hits@50 & MRR \\
    \midrule
    MLP & 0.0046 ± 0.0000 & 0.1927 ± 0.0129 & 0.2904 ± 0.0013 \\
    CN & 0.2765 ± 0.0000 & 	0.5006 ± 0.0000 & 0.5147 ± 0.0000 \\
    Node2vec & 0.2226 ± 0.0083 & 0.4888 ± 0.0054 & 0.6141 ± 0.0011 \\
    GCN & 0.1867 ± 0.0132 & 0.4475 ± 0.0107 & 0.8474 ± 0.0021 \\
    GraphSAGE & 0.1655 ± 0.0240 & 0.4810 ± 0.0081 & 0.8260 ± 0.0036 \\
    SEAL & 0.4880 ± 0.0316 & 0.5371 ± 0.0047 & 0.8767 ± 0.0032 \\
    DE-GNN & 0.3648 ± 0.0378 & 0.5303 ± 0.0050 & 0.6030 ± 0.0061 \\
    SEG & \textbf{0.5059} ± 0.0239 & \textbf{0.5535} ± 0.0039 & \textbf{0.8773} ± 0.0030 \\
    \bottomrule
  \end{tabular}
\end{table*}

\subsubsection{Datasets.} 
We test the proposed SEG for link prediction on the Open Graph Benchmark (OGB)~\cite{hu2020open} datasets. OGB datasets are open-sourced large-scale datasets which cover a diverse range of domains, ranging from social networks, biological networks to molecular graphs. OGB provides standard evaluation metrics to fairly compare the performance of each algorithm. We use three link prediction datasets in OGB including ogbl-ppa~\cite{szklarczyk2019string}, ogbl-collab
~\cite{wang2020microsoft} and ogbl-citation2~\cite{wang2020microsoft}.
The ogbl-ppa dataset is an undirected, unweighted protein-protein association graph with 576,289 nodes and 30,326,273 edges. Each node contains a 58-dimensional embedding that indicates the species that the corresponding protein comes from.
The ogbl-collab dataset is an undirected graph, representing a collaboration network between authors. It contains 235,868 nodes and 1,285,465 edges. Nodes represent authors and edges
indicate the collaboration between authors. Each node contains 128-dimensional features, generated
by averaging the word embeddings of authors' papers.
The ogbl-citation2 dataset is a directed citation graph between
a subset of papers with 2,927,963 nods and 30,561,187 edges. Nodes represent papers and each edge indicates that one paper cites another. Each node with pre-trained 128-dimensional Word2vec embedding with its title and abstract. The detailed statistics of the datasets are summarized in Table~\ref{tab:data}.

\subsubsection{Baselines.} 
We compare SEG with representative and
state-of-the-art link prediction algorithms, which includes:
\begin{itemize}
    \item MLP~\cite{pal1992multilayer} is a basic neural network with node features directly as input. MLP is used to verify whether the original node feature is enough to predict links.
    \item CN~\cite{newman2001clustering} is a common approach for link prediction that computes the number of common neighbors between two nodes as the similarity score of a link between the nodes. Nodes with higher scoring are more likely to have a link.
    \item Node2vec~\cite{grover2016node2vec} is a classical graph embedding method that encodes nodes into low dimensional vectors. Then a MLP predictor is applied that uses the combination of the original node features and the Node2vec output vectors as input.
    \item GCN~\cite{kipf2016semi} is a well-known graph neural network which defines graph convolution via spectral analysis.
    \item GraphSAGE~\cite{hamilton2017inductive} is a widely used graph neural network that uses sample and aggregation tricks to support inductive learning for unseen nodes.
    \item SEAL~\cite{zhang2018link} is a link prediction method that learns link representations from labeled subgraphs via the Deep Graph Convolutional Neural Network (DGCNN)~\cite{zhang2018end}.
    \item DE-GNN~\cite{li2020distance} is a GNN model which uses shortest-path-distances node labeling instead of DRNL in SEAL. 
\end{itemize}

\subsubsection{Evaluation Metric.} 
In this paper, we adopt metrics that include top-K hit rate (Hits@K) and Mean Reciprocal Rank (MRR). 
Hits@K counts the ratio of positive edges that are ranked at the K-th place or above against all the negative edges. Hits@K is more challenging than well-known AUC because the model needs to rank positive edges above almost all negative edges.
MRR computes the reciprocal rank of the true target edge among the 1,000 negative candidates for each source node and then averages over all these source nodes.
Both metrics are higher the better.

\subsubsection{Settings and Hyperparameters.} 
We follow the original OGB dataset splitting, the training/validation/testing set is split as follows, for ogbl-ppa is 70/20/10, for ogbl-collab is 92/4/4, and 98/1/1 for ogbl-citation2. 
GCN and GraphSAGE have 3 layers with 256 hidden dimensions and two fully-connected layers, each contains 256 hidden units that are applied as the predictor of link existence. 
For SEAL and DE-GNN, the labeled structural features are concatenated with the node features as input of the GNN model which has 3 layers with 32 hidden units. For a fair comparison with SEAL, we also use a 3 layers with 32 hidden units DGCNN as base model in SEG. DE-GNN uses a 3 layers GCN as its base model.
DE-GNN, SEAL and SEG use a 2 layers MLP predictor with 128 units to output the probability of link.
For structure encoder in SEG, we use a single layer GCN and 3 layers MLP to encode path features and then use another 2 layers MLP to output the link scores.
In the training, for ogbl-ppa we use only 5$\%$ of the training edges as positive samples, for ogbl-collab we use 10$\%$ of the data, and for ogbl-citation2 we use 2$\%$ of the data, which is the same as the setting of SEAL. We set $\lambda$ to 4 for all the three datasets to prevent overfitting.
The negative link (edge consisting of two nodes that are not connected) is sampled randomly from the training set. We report the averaged results over 10 times of running.

\begin{table}[htb]
  \caption{Hits@50 results for ogbl-collab with validation data as input.}
  \label{tab:collab_val}
  \begin{tabular}{l|c}
    \toprule
	Method & ogbl-collab \\
    \midrule
    CN & 0.6137 ± 0.0000 \\
    GCN & 0.4714 ± 0.0145 \\
    GraphSAGE & 0.5463 ± 0.0112	\\
    SEAL & 0.6364 ± 0.0071 \\
    DE-GNN & 0.6440 ± 0.0028 \\
    SEG & \textbf{0.6445} ± 0.0036 \\
    \bottomrule
  \end{tabular}
\end{table}

\subsection{Basic Results}

We conduct experiments on three public OGB datasets to answer the first question.
It is worth mentioning that the SEAL method has achieved the best performance on these three datasets so far. Experimental results on three OGB datasets are reported in Table~\ref{tab:ogb}. Results show that the proposed method SEG beats SEAL and achieves the best performance on all three datasets.

On both ogbl-ppa and ogbl-collab datasets, SEG achieves 3$\%$ improvement over the SEAL method. The topology-based method CN achieves higher Hits@K than the Node2vec and GNNs which indicates that the structural information is more important than the original node features in these graphs. 
Besides, the MLP baseline performs extremely poorly on all three datasets, which shows that the attributes of the nodes in these datasets do not accurately reflect the interaction between the nodes. 

As a matter of fact, in many real-world graphs, it is difficult to design high-quality semantic features for target tasks. For these graphs, it is better to effectively use the structural information from graph itself.

For larger scale ogbl-citation2, SEG still outperforms all baselines, which shows the effectiveness of the proposed method. DE-GNN, SEAL and SEG achieve a huge improvement over other GNN models on both ogbl-ppa and ogbl-collab, which illustrates the effectiveness of adding structural information to GNNs. 
Note that validation edges are also allowed to be included in training~\cite{hu2020open}. So we conduct experiments to use the validation edges as training as well. The results in Table~\ref{tab:collab_val} show that SEG is still better than all other methods.

\subsection{Ablation studies}

We next attempt to answer the second and third questions raised at the beginning of this section. 
We first only use structure encoding part in SEG denoted as SEG-SE, and compare the SEG-SE with SEAL-PL which uses path labeling instead of DRNL in SEAL on the ogbl-collab dataset.
The result in Table~\ref{tab:collab} shows that SEG-SE achieves 0.5503 Hits@50, which is better than SEAL-PL. It demonstrates the effectiveness of our structure encoding. We then remove the final predictor MLP and $s_{structure}$ of structure encoding in SEG, denoted as SEG-GNN, and compare it with SEG-SE and SEG. As shown in Table~\ref{tab:collab}, within these three methods, SEG achieves the best results, which illustrates the benefits of joint training.
By joint training, structural embedding can better represent the original structural features.

It is worth noting that the result of SEG-GNN is still better than SEAL, which means our structure encoding method is more effective than DRNL of SEAL~\cite{zhang2018link}.
\begin{table}[htb]
  \caption{Hits@50 results for ogbl-collab}
  \label{tab:collab}
  \begin{tabular}{l|c}
    \toprule
	Method & ogbl-collab \\
    \midrule
    SEAL & 0.5371 ± 0.0047 \\
    SEAL-PL & 0.5393 ± 0.0052 \\
    SEG-GNN & 0.5465 ± 0.0021 \\
    SEG-DRNL & 0.5476 ± 0.0051 \\
    SEG-SE & 0.5503 ± 0.0033 \\
    SEG & \textbf{0.5535} ± 0.0039 \\
    \bottomrule
  \end{tabular}
\end{table}
This also proves our proposition, which is that the structural information and the original node features are in different spaces and cannot be directly and simply concatenated. Instead, we use a feature fusion module to map these two features to the same space.

To further address the difference between our Path Labeling(PL) and DRNL of SEAL, we conduct an experiment that uses SEAL's DRNL features as the structural features in the SEG framework, denoted as SEG-DRNL in Table~\ref{tab:collab}. The result shows that SEG outperforms SEG-DRNL. 
In addition, we replace the DRNL in SEAL with PL named SEAL-PL, and compare it with the original SEAL. The result shows that SEAL-PL is better than original SEAL. It means under the same training framework and settings, the PL are more effective than DRNL.
Our analysis of the label values reveals that for the 1-hop subgraph, DRNL has a total of 17 different values of node labels, much more than the 4 of PL. 
This indicates that more label values introduce noise, so we do not distinguish between labels of nodes of the same path and place a limit on the length of the path to further prevent overfitting.
Besides, the Hits@50 of SEG-DRNL is higher than Hits@50 of SEAL on ogbl-collab. It also illustrates the effectiveness of our SEG framework.

\subsection{Caveats}
There is an ogbl-ddi graph~\cite{ddi} in OGB link prediction datasets which is much denser than the other datasets introduced above. The ogbl-ddi dataset is an undirected graph. Each node represents a drug and the edge represents the interaction between drugs. It has 1,334,889 edges but only has 4,267 nodes. The average degree of the nodes in this graph is 500.5 and the density is 14.67$\%$, which is much larger than the other three graphs (see statistics of all datasets in Table~\ref{tab:degree}). Nodes in ogbl-ddi do not contain any feature, so we exclude the MLP method in comparison. All learning-based methods use free-parameter node embeddings as the input. In the SEG method, we only use GNN and exclude structure encoder.

\begin{table}
  \caption{Statistics of the OGB link prediction datasets}
  \label{tab:degree}
  \begin{tabular}{l|cc}
    \toprule
	Dataset & Average Node Degree & Density \\
    \midrule
    ogbl-ppa & 73.7  & 0.018$\%$ \\
    ogbl-collab & 8.2 & 0.0046$\%$ \\
    ogbl-citation2 & 20.7 & 0.00036$\%$ \\
    ogbl-ddi & 500.5 & 14.67$\%$ \\
    \bottomrule
  \end{tabular}
\end{table}

\begin{table}
  \caption{Hits@20 results for ogbl-ddi}
  \label{tab:ddi}
  \begin{tabular}{l|c}
    \toprule
	Method & ogbl-ddi \\
    \midrule
    CN & 0.1773 ± 0.0000 \\
    Node2vec & 0.2326 ± 0.0209 \\
    GCN & 0.3707 ± 0.0507 \\
    GraphSAGE & \textbf{0.5390} ± 0.0474 \\
    SEAL & 0.3056 ± 0.0386 \\
    DE-GNN & 0.2663 ± 0.0682 \\
    SEG & 0.3190 ± 0.0314 \\
    \bottomrule
  \end{tabular}
\end{table}

Table~\ref{tab:ddi} shows experimental results.
The results on ogbl-ddi indicate that DE-GNN, SEAL and SEG, all three GNN methods using structural features do not perform well on dense graphs. Although SEG still beats SEAL, it falls behind ordinary GNNs methods like GCN and GraphSAGE. This result shows the local structural information is no longer distinguishable in this dense graph. Note that the topology-based CN method has the worst performance, which also indicates that on this dense graph it is hard to learn meaningful structural patterns.
However, most large-scale real-world graphs (e.g., citation graphs, internet graphs, etc.) are power-law~\cite{clauset2009power, faloutsos2011power}, so local graph structure is useful for these graphs.
For dense graph, we need to consider a larger range of global structural features, this is left to our future work.

\section{Conclusion} \label{sec:conclusion}
In this paper, We first introduce structure encoding method to capture the target nodes' surrounding topology information for link prediction. We further propose SEG, a novel GNN framework that fuses node features with graph structural information to enhance the power of GNN for link prediction. 
Experimental results on OGB link prediction datasets demonstrate the effectiveness of the proposed framework. In the future, we will extend our method for heterogeneous graphs by considering different types of edges.


\bibliographystyle{ACM-Reference-Format}
\bibliography{main}

\end{document}